\documentclass{article}

     \usepackage[preprint]{neurips_data_2024}
\usepackage[utf8]{inputenc} % allow utf-8 input
\usepackage[T1]{fontenc}    % use 8-bit T1 fonts
\usepackage{hyperref}       % hyperlinks
\usepackage{url}            % simple URL typesetting
\usepackage{booktabs}       % professional-quality tables
\usepackage{amsfonts}       % blackboard math symbols
\usepackage{nicefrac}       % compact symbols for 1/2, etc.
\usepackage{microtype}      % microtypography
\usepackage{xcolor}         % colors

\usepackage{amsmath,amssymb}
\usepackage{graphicx}
\usepackage{listings}

\title{MathOdyssey: Benchmarking Mathematical Problem-Solving Skills in Large Language Models Using Odyssey Math Data}

\author{Meng Fang$^{\,1}$\hspace{5pt} Xiangpeng Wan$^{\,2}$\hspace{5pt} Fei Lu$^{\,3}$\hspace{5pt} Fei Xing$^{\,4}$\hspace{5pt} Kai Zou$^{\,2,5}$\\
  $^{1\,}$University of Liverpool\hspace{5pt} $^{2\,}$NetMind.AI\\
  	$^{3\,}$Johns Hopkins University\hspace{5pt} $^{4\,}$Mathematica Policy Research \\
   $^{5\,}$AGI Odyssey \\
  \texttt{hello@agiodyssey.org} \\
}

\begin{document}

\maketitle

\begin{abstract}
Large language models (LLMs) have significantly advanced natural language understanding and demonstrated strong problem-solving abilities. Despite these successes, most LLMs still struggle with solving mathematical problems due to the intricate reasoning required. This paper investigates the mathematical problem-solving capabilities of LLMs using the newly developed ``MathOdyssey'' dataset. The dataset includes diverse mathematical problems at high school and university levels, created by experts from notable institutions to rigorously test LLMs in advanced problem-solving scenarios and cover a wider range of subject areas. By providing the MathOdyssey dataset as a resource to the AI community, we aim to contribute to the understanding and improvement of AI capabilities in complex mathematical problem-solving.
We conduct benchmarking on open-source models, such as Llama-3 and DBRX-Instruct, and closed-source models from the GPT series and Gemini models. Our results indicate that while LLMs perform well on routine and moderately difficult tasks, they face significant challenges with Olympiad-level problems and complex university-level questions. Our analysis shows a narrowing performance gap between open-source and closed-source models, yet substantial challenges remain, particularly with the most demanding problems. This study highlights the ongoing need for research to enhance the mathematical reasoning of LLMs. 
The dataset, results, and code are publicly available.\footnote{\url{https://mathodyssey.github.io/}}
\end{abstract}

\section{Introduction}

Large language models (LLMs) have demonstrated exceptional proficiency in mastering human language and handling mathematical problems, including typical routine math problems \citep{achiam2023gpt,touvron2023llama,reid2024gemini}. In recent years, several benchmarks related to mathematics have been proposed, such as the GSM8K dataset \citep{cobbe2021gsm8k}, the MATH dataset \citep{mathdataset2021} and so on. Recent LLMs and prompting approaches have addressed these problems with notable success \citep{achiam2023gpt,touvron2023llama,touvron2023llama2}. For instance, GPT-4, using advanced prompting techniques \citep{achiam2023gpt}, has achieved more than a 90\% success rate on GSM8K and 80\% on MATH. These achievements indicate that LLMs possess remarkable capabilities in mathematical reasoning.

The quest to improve LLMs' mathematical problem-solving abilities is not just a demonstration of technological advancement but a crucial step toward developing more general and capable artificial intelligence systems. On the one hand, this endeavor requires datasets that accurately measure and challenge the AI's mathematical reasoning beyond basic problems. Although their performance is high on datasets like GSM8K \citep{cobbe2021gsm8k}, it remains uncertain how well they handle more complex mathematical challenges, such as those found in university-level courses and competitive high school mathematics. Performance may diminish significantly in these areas. This gap highlights the ongoing need for enhanced mathematical reasoning capabilities in AI, a critical area for assessing cognitive abilities akin to human intelligence.
Moreover, a significant obstacle is that many existing datasets might have been included in the training phases of these models, potentially skewing performance metrics. Prominent examples include STEM-Q \citep{stemq-drori2023}, GSM8K \citep{cobbe2021gsm8k}, and the MATH dataset \citep{mathdataset2021}, which may no longer provide a true test of an LLM's mathematical capabilities.
On the other hand, high-quality, expert-crafted original problems are scarce. For instance, a study by OpenAI \citep{davis2023testing} included only 105 such problems in high school and university-level science and math.

\begin{table*}[t] % This will allow the table to span both columns
\centering
\label{tab:examples}
%\scriptsize % Sets a smaller font size to help fit the table
\setlength{\tabcolsep}{3pt} % Reduce padding between columns
\begin{tabular}{p{13cm}}
\toprule \vspace{-3mm}
\begin{center} \textbf{Olympiad-level}\end{center}
\textbf{Problem:} Let $S=\left\{ 1,2,\cdots 2024 \right\}$, if the set of any $n$ pairwise prime numbers in $S$ has at least one prime number, the minimum value of $n$ is \underline{\hspace{2cm}}.\\
\textbf{Answer:} $16$. \\
\textbf{Reasoning:} Taking the 15 numbers $1, 2^2, 3^2, ..., 43^2$. They violate the
condition. Furthermore, since S does not contain any non-prime numbers with a minimum prime factor of at least 47 (because $47^2 > 2024$). Set 1 aside, there are only 14 types of non-prime numbers in S, classified by its minimum prime factor. Applying the Pigeonhole Principle, we conclude that n = 16. \\ \hline 
\vspace{-3mm}
\begin{center} \textbf{High School} \end{center}
\textbf{Problem:} What are the solutions of the quadratic equation $15x^2=2x+8$.
\begin{align*}
\text{ A) }  \, \left\{-\frac{4}{3},-\frac{3}{2}\right\}  \,\,
~~\text{ B) } \, \left\{-\frac{4}{5},\frac{2}{3}\right\}   \,\,
~~\text{ C) } \, \left\{-\frac{3}{2},\frac{4}{5}\right\}   \,\,
~~\text{ D) } \,  \left\{-\frac{2}{3},\frac{4}{5}\right\} 
\end{align*} 
\textbf{Answer:} $D$ \\
\textbf{Reasoning:}
First move all terms to one side: $15x^2-2x-8=0$. Then factor into $(5x-4)(3x+2)=0$. Setting $5x-4$ to zero results in a solution of $x = \frac{4}{5}$ and setting $3x+2$ to zero results in a solution of $x = -\frac{2}{3}$. \\ \hline 
\vspace{-3mm}
\begin{center} \textbf{University-level} \end{center}
\textbf{Problem:} Find the limit $$\lim\limits_{x\to 1}\frac{f(2x^2+x-3)-f(0)}{x-1}$$ given $f'(1)=2$ and $f'(0)=-1$.
\\
\textbf{Answer:}  $-5$.\\
\textbf{Reasoning:} 
Let $g(x)=2x^2+x-3$. Since $g(1)=0$, the desired limit equals $\lim\limits_{x\to 1}\frac{f(g(x))-f(g(1))}{x-1}$. By the definition of the derivative and the chain rule and noting that $g'(1)=5$, we have
\[
\lim\limits_{x\to 1}\frac{f(g(x))-f(g(1))}{x-1}=f'(g(1))g'(1)=f'(0)g'(1)=(-1)(5)=-5.
\]\\
\bottomrule
\end{tabular}
\caption{MathOdyssey dataset examples. We demonstrate three distinct levels to challenge various aspects of mathematical knowledge: Olympiad-level, High School, and University-level mathematics. Each example consists of three parts: the problem, the answer, and the reasoning. Note that both GPT-4 Turbo and Llama-3-70B are unable to solve the first Olympiad-level example. See Appendix A for the LLMs' solutions. 
}
\end{table*}

To directly address these challenges, we introduce the ``MathOdyssey'' dataset, a rigorously curated collection of 387 mathematical problems for evaluating the general mathematical capacities of LLMs. See examples in Table \ref{tab:examples}. The MathOdyssey dataset is developed by the GAIC Math organization and features a spectrum of questions from Olympiad-level competitions, advanced high school curricula, and university-level mathematics. 
Mathematics professionals, including high-school educators, researchers, and university professors, crafted these problems under the invitation of the GAIC Math organization. Their involvement ensures the dataset not only supports advanced AGI research but also fosters necessary interdisciplinary collaboration. 

Furthermore, we open-source the MathOdyssey dataset to facilitate its use in evaluating other LLMs. The dataset has not been used for training by LLMs. We explore its utility in benchmarking the advanced mathematical reasoning abilities of LLMs. By ensuring the originality and confidentiality of the questions, we maintain the integrity and fairness of the assessments, providing a reliable tool for advancing research into artificial general intelligence.

Our contributions are as follows:
\begin{itemize}
\item We introduce a new mathematical challenge that provides different levels of mathematical problems and covers a wider range of subject areas.
%\item We introduce a new mathematical challenge. Unlike the AI-MO challenge, we provide different level mathematical problems and cover a wider range of subject areas.

\item We open source the MathOdyssey benchmark dataset, a meticulously curated collection of mathematical problems spanning various domains and levels, complete with natural language solutions. This dataset is specifically designed to probe the reasoning abilities of LLMs, offering a unique tool for assessing AI performance in complex mathematical reasoning. Each question has an objective answer serving as ‘ground-truth’, allowing for objective evaluation on the LLM outputs. In particular, the Open-Answer problems emphasize the importance of detailed reasoning and solution.

\item We conduct a comprehensive benchmark analysis using our dataset on both open-source and closed-source LLMs. Our findings reveal that while closed-source models currently lead, open-source models are rapidly catching up, highlighting the competitive landscape of LLM capabilities in mathematical problem-solving.
\end{itemize}

\section{Related Work}
%\mf{to revise} \citep{achiam2023gpt}
%In the realm of mathematical problem-solving using open-source large language models (LLMs), significant strides have been made. Innovations such as MathBert \cite{mathbert2021} and Lean-Dojo \cite{leandojo-yang2023} demonstrate the fine-tuning of these models to tackle specific types of mathematical challenges effectively. Meanwhile, MathGLM \cite{mathGLM2023} has been tailored to excel in elementary arithmetic operations, and WizardMath \cite{luo2023wizardmath} employs a reinforced evolution-instruct approach to enhance LLMs' mathematical competencies. Additionally, Math-Coder \cite{wang2023mathcoder} integrates the Llama model with programming capabilities, significantly enhancing its mathematical problem-solving prowess. MathGPT\footnote{https://www.mathgpt.com/} also stands out by demonstrating impressive capabilities in solving K-12 mathematical problems, underscoring the potential of open-source models to rival their closed-source counterparts.

\paragraph{Large Language Models for Mathematics.}
Applying large language models (LLMs) to mathematical problems has led to significant strides, though solving such problems remains challenging due to the need for highly complex and symbolic multi-step reasoning capabilities. Both GPT-3.5 and GPT-4 \citep{achiam2023gpt} have shown promising reasoning abilities for complex mathematical tasks, such as those in the MATH dataset \citep{hendrycksmath2021}. However, the performance of open-source models, like Llama-1 and Llama-2 \citep{touvron2023llama,touvron2023llama2}, is still far from satisfactory in this domain. To enhance the mathematical problem-solving abilities of LLMs, prompt-based methods have also been developed \citep{wei2022chain, wang2022self, zhou2022least}. These methods aim to improve reasoning and accuracy by guiding the models through structured prompts that help in breaking down complex problems into manageable steps.

\paragraph{Mathematical Evaluation for Large Language Models.}
Evaluating the mathematical capacity of large language models (LLMs) is crucial. Benchmarks such as GSM8K \citep{cobbe2021gsm8k}, which targets middle-school level mathematics, and MATH \citep{hendrycksmath2021}, which focuses on high-school math competitions, have been widely used. For university-level problems, datasets like ProofNet \citep{azerbayev2023ProofNet} and OCWCourses \citep{lewkowycz2022solving} are prominent. Additionally, MiniF2F \citep{zheng2022_MiniF2FCrosssystem} and AlphaGeometry \citep{trinh2024_SolvingOlympiad} provide Olympiad-level problems, while the SAT dataset \citep{azerbayev2023_LlemmaOpen} includes problems from the College Board SAT examination.
These datasets have limitations, particularly at the undergraduate level and above, where they fall short in addressing graduate-level and competition-level difficulties \citep{frieder2024mathematical}. To address this gap, we introduce the MathOdyssey dataset, a diverse collection of mathematical problems designed to serve as a rigorous benchmark for assessing both open-source and closed-source models. Table \ref{tab:testing_datasets} highlights the properties of MathOdyssey compared to relevant benchmarks, emphasizing the different levels and the diversity of subject areas and question types in our benchmark. This dataset spans a spectrum of difficulty levels, from high school to advanced university mathematics, highlighting the evolving capabilities and ongoing challenges in LLM mathematical problem-solving.

\begin{table}[t]
\centering
\small
\begin{tabular}{|l |c | l | l|}
\hline
\textbf{Dataset} & \textbf{Year} & \textbf{Description}    &   \textbf{\# of Test}                         \\ \hline
GSM8k \citep{cobbe2021gsm8k}&  2021       &     8.5k middle-school level math word problems & 1k \\ \hline
MATH \citep{mathdataset2021}& 2021         & 12.5k high-school math competitions & 5k  \\ \hline
OCWCourses \citep{lewkowycz2022solving}& 2022     & University-level, MIT's OpenCourseWare&  272     \\ \hline
MiniF2F \citep{zheng2022_MiniF2FCrosssystem}&2023 &  Olympiad-level   &  488  \\ \hline
SAT \citep{azerbayev2023_LlemmaOpen} &  2023& Figureless questions from SAT  & 32 \\ \hline
ProofNet \citep{azerbayev2023ProofNet} & 2023  & University-level,  proofs &  371  \\ \hline
AlphaGeometry \citep{trinh2024_SolvingOlympiad}& 2024 & Olympiad Geometry only & 30 \\ \hline
MathOdyssey (this work) & 2024     & High School, University-level, Olympiad-level& 387     \\ \hline
\end{tabular}
\caption{Comparison of existing evaluation datasets for testing AI in mathematics. These datasets are limited, especially in the availability of high-quality, expert-crafted original problems with varying difficulty levels.} 
\label{tab:testing_datasets}
\end{table}

%Contrastingly, large closed-source models like GPT-4 and Claude excel in non-university-level mathematical queries, as evidenced by their performance on datasets such as GSM8k \cite{cobbe2021gsm8k} and Math \cite{mathdataset2021}. However, their limitations become apparent at the undergraduate level and above, particularly with graduate-level and competition-level difficulties \cite{frieder2024mathematical}. Building on this foundation, our study introduces the MathOdyssey Dataset, a diverse collection of mathematical problems designed to serve as a rigorous benchmark for assessing both open-source and closed-source models. This dataset fills a critical gap by providing challenges across a spectrum of difficulty levels, from high school to advanced university mathematics. Furthermore, our analysis reveals that the open-sourced models are approaching the capabilities of an earlier version of GPT-4 and close-sourced model, GPT-4 series, exhibit stronger performance in high school and university-level math, highlighting ongoing advancements in their development. This research not only contributes a valuable tool for the AI community but also highlights the evolving capabilities and ongoing challenges in the field of AI-driven mathematical problem-solving.

\section{MathOdyssey}
To evaluate the mathematical reasoning abilities of LLMs, we create the MathOdyssey dataset, a rigorously curated collection designed by professionals from both universities and high schools. 
To ensure comprehensive evaluation and promote transparency, we have made the entire MathOdyssey dataset and benchmarking code publicly available.
This allows other researchers to replicate our study, compare methods, and explore new approaches using the dataset.

\subsection{Data Collection}
\paragraph{Design Principle.}
The motivation behind the design of the MathOdyssey dataset is to establish a new benchmark representing the pinnacle of human intellectual achievement, encouraging researchers to push the boundaries of LLMs' mathematical reasoning capabilities. To realize this vision, we have curated challenges that epitomize comprehensive levels of math problems. Specifically, our benchmark includes:
\begin{itemize}
    \item Inclusion of diverse levels of math problems: Ensuring a comprehensive understanding and catering to various proficiency levels promotes a well-rounded mastery of mathematical concepts and problem-solving skills. This dataset offers a range of problems, starting from basic concepts and gradually increasing in difficulty to cover advanced topics. This allows for a thorough evaluation of AI capabilities across various levels of high school and university mathematics.

\item Inclusion of different subject area problems: Enhancing LLMs' mathematical proficiency by exposing them to a wide range of concepts and techniques, from foundational arithmetic to advanced topics such as algebra, number theory, geometry, combinatorics, and calculus. These diverse subject areas help identify LLMs' strengths and areas for improvement, encouraging the development of critical mathematical reasoning, problem-solving skills, and a deeper appreciation for the interconnected nature of mathematics. By integrating various mathematical disciplines, researchers can create a more engaging and comprehensive learning environment that prepares LLMs for complex real-world challenges in mathematics.

\item Provision of objective answers and detailed solutions: The objective answers serve as ‘ground-truth’, allowing for objective evaluation of the LLM outputs. In particular, the Open-Answer problems emphasize the importance of detailed reasoning and solution.
Given the varying difficulty and subject areas of these problems, which may exceed comprehension without a specialized background in mathematics, each problem is accompanied by expertly crafted solutions detailing the reasoning steps involved. These solutions are useful for evaluation and can enhance the assessment of LLMs' reasoning processes.
\end{itemize}

\paragraph{Human professionals.} The dataset was created by human professionals to ensure high quality. Experts developed a wide range of mathematical problems for the MathOdyssey dataset, featuring a spectrum of questions from Olympiad-level competitions, advanced high school curricula, and university-level mathematics. Mathematics professionals, including high-school educators, university professors, and researchers,  crafted these problems. Their involvement ensures the dataset not only supports advanced AGI research but also fosters necessary interdisciplinary collaboration.

A typical problem in the MathOdyssey dataset comprises three components: the problem, the answer, and the reasoning, as detailed in Table \ref{tab:examples}. The problems are original and not sourced from previous datasets or textbooks. Each problem is accompanied by an answer and a detailed solution that explains the reasoning process used to derive the answer. After creation, the problems undergo independent review by a separate team of researchers with expertise in mathematics. This team assesses the problems and their solutions, eliminating any ambiguous or redundant responses to enhance the set's validity and reliability. This rigorous process guarantees the quality and dependability of the final problem set.

\subsection{Dataset Analysis}

\begin{figure}[t!]
	\centerline{\includegraphics[width=\textwidth]{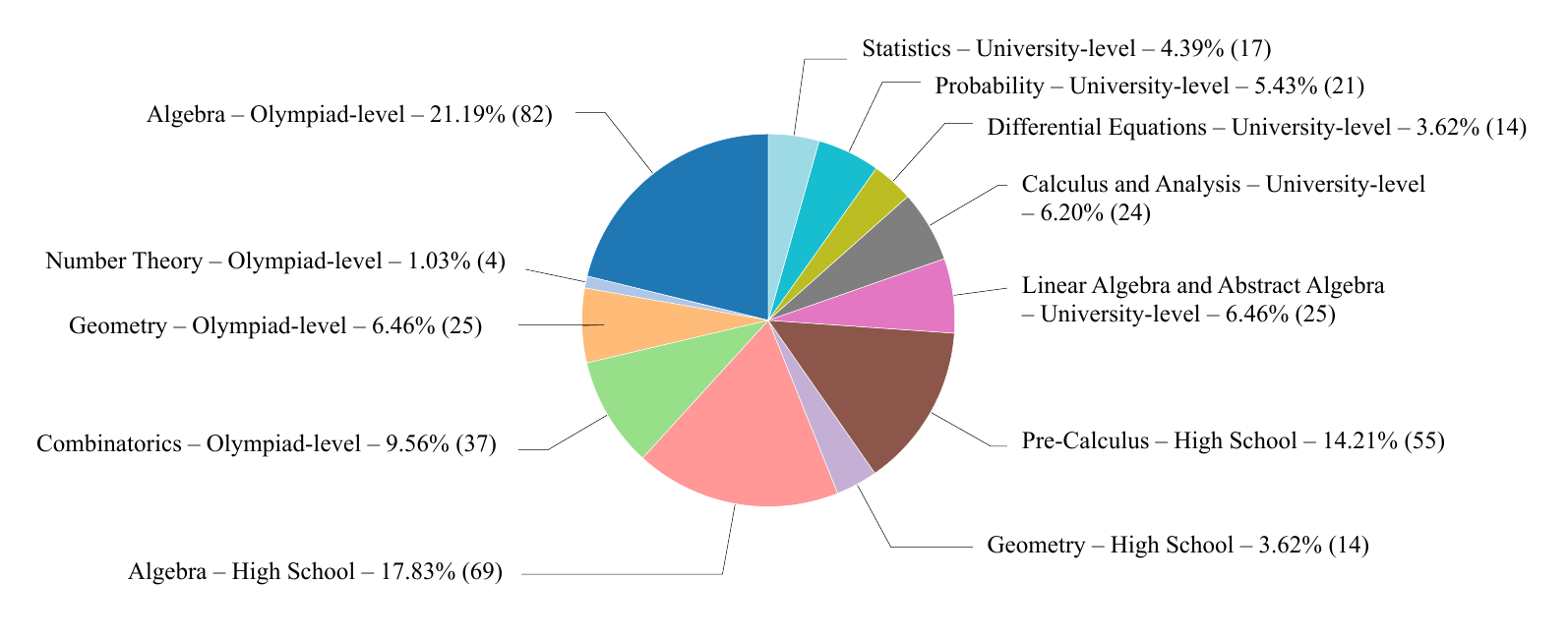}}
	\caption{Mathematical problems across educational levels. We curate and categorize problems by difficulty and subject area. 
 }\label{fig:data}
\end{figure}

To understand the properties of the MathOdyssey dataset, we analyze the questions and answers. Specifically, we explore (i) the difficulty of questions based on the type of reasoning required to answer them, (ii) the subject areas of the problems, and (iii) the diversity of answer types.

\paragraph{Difficulty of questions.}
In the MathOdyssey dataset, each category is designed to evaluate different facets of mathematical reasoning and problem-solving capabilities, ranging from fundamental high school concepts to complex university-level theories, as summarized in Figure~\ref{fig:data}. This diverse dataset is structured into three distinct levels to challenge various aspects of mathematical knowledge:
\begin{itemize}
\item \textbf{Olympiad-level}: It tests advanced problem-solving skills with questions in Algebra, Number Theory, Geometry, and Combinatorics.
\item \textbf{High School}: Broadening the scope, this category includes problems in Algebra, Geometry, and Pre-Calculus, covering a comprehensive range of high school math concepts.
\item \textbf{University-level}: Catering to higher education, this segment offers challenges in Linear and Abstract Algebra, Calculus and Analysis, Differential Equations, Probability, and Statistics, suitable for university students.
\end{itemize}

The MathOdyssey dataset categorizes mathematical problems across different educational levels, helping to understand the distribution and scope of problems included in the dataset. 
For Olympiad-level Competition, the categories and their respective percentages are Algebra (21.19\%), Number Theory (1.03\%), Geometry (6.46\%), and Combinatorics (9.56\%), totaling 38.24\%.
For High School Mathematics, the categories are Algebra (17.83\%), Geometry (3.62\%), and Pre-Calculus (14.21\%), totaling 35.66\%.
For University-level, the categories are Linear and Abstract Algebra (6.46\%), Calculus and Analysis (6.20\%), Differential Equations (3.62\%), Probability (5.43\%), and Statistics (4.39\%), totaling 26.10\%.
Three subject areas, Differential Equations, Probability, and Statistics, only appear at the University level.

\paragraph{Subject areas of the problems.}
The problems encompass a wide range of topics, including Algebra, Number Theory, Geometry, Combinatorics, Pre-Calculus, Linear and Abstract Algebra, Calculus and Analysis, Differential Equations, Probability, and Statistics, as shown in Figure~\ref{fig:data}. The MathOdyssey dataset encompasses a wide range of subject areas, providing a comprehensive testing ground for the mathematical reasoning and problem-solving capabilities of large language models (LLMs). Algebra problems constitute 21.19\% from Olympiad-level Competition and 17.83\% from High School Mathematics, making them the most represented areas in the dataset. In contrast, Number Theory problems, with only 1.03\% from Olympiad-level Competition, have the lowest representation. Pre-Calculus problems, accounting for 14.21\% of High School Mathematics, play a significant role in preparing students for more advanced calculus topics. Other subject areas, including Calculus and Analysis, Linear and Abstract Algebra, Differential Equations, Probability, and Statistics, each contribute around 4\% to 8\% to the dataset. See Appendix B for examples that help better understand the reasoning required to answer the questions.
%For each math problem, experts provided the answers as well as detailed solutions or reasoning processes from a human perspective.

%The Global Artificial Intelligence Championship (GAIC) Math 2024 features a meticulously curated collection of 387 mathematical problems, designed by professional problem writers from both universities and high schools. 

\begin{figure}[t!]
	\centerline{\includegraphics[width=\textwidth]{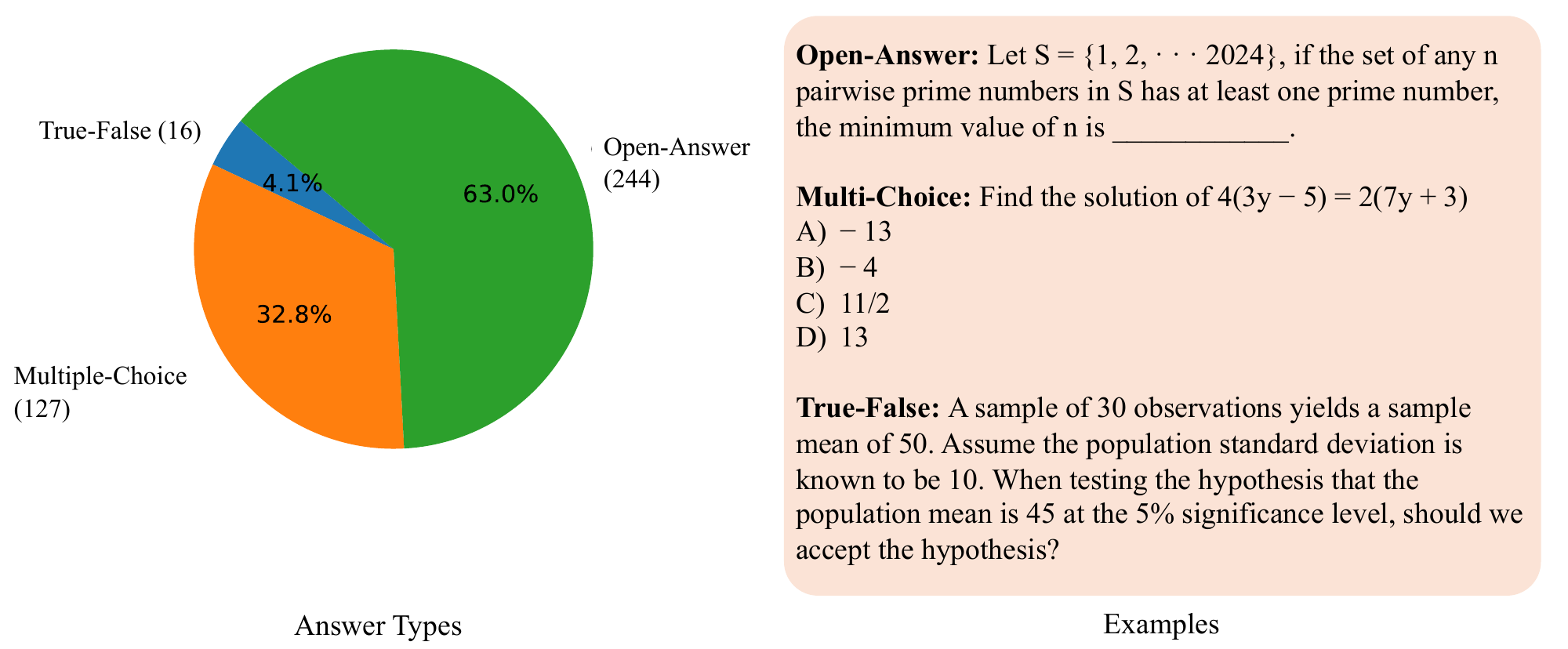}}
	\caption{There are three answer-types: True-False questions, Multiple-Choice questions and Open-Answer questions.}\label{fig:answer-types}
\end{figure}

\paragraph{Diversity of answer types.} 
%We utilize answer types commonly found in standard math exams and competitions, including True-false questions, Multiple-choice questions, and Open-answer questions, as shown in Figure \ref{fig:answer-types}. 
The MathOdyssey dataset includes a variety of answer types, providing a comprehensive assessment of the mathematical reasoning and problem-solving capabilities of large language models (LLMs). The distribution of answer types is shown in Figure \ref{fig:answer-types}, and it is categorized into three main types: True-False questions, Multiple-Choice questions, and Open-Answer questions.
The distribution of answer types in the MathOdyssey dataset is designed to provide a well-rounded evaluation of LLMs' mathematical capabilities. With 63.0\% of the questions being open-answer, the dataset emphasizes the importance of detailed reasoning and solution generation. Multiple-choice questions, making up 32.8\%, help assess the models' ability to choose correct answers from given options, while true-false questions, at 4.1\%, provide a quick check of fundamental understanding.
This diverse mix of answer types ensures that LLMs are tested on various aspects of mathematical problem-solving, from basic validation to complex reasoning and solution generation, requiring an understanding of the concepts.

\section{Experiments}

Our goal is to provide a comprehensive standardized dataset to evaluate LLMs on mathematical reasoning. By comparing different models, our benchmarks highlight their strengths and weaknesses.

\subsection{Models}
%Our dataset was used to test a range of state-of-the-art LLMs, both open-sourced and closed-sourced, providing a robust benchmark for assessing their mathematical problem-solving capabilities.
We evaluate both open-source and closed-source LLMs. The models tested include GPT-4 Turbo, GPT-4 \citep{achiam2023gpt}, GPT-3.5 Turbo, Gemini models \citep{reid2024gemini}, Claude 3 \citep{anthropic}, Llama-3-70B, and DBRX-Instruct \citep{themosaic2024intro}. All models are tested using chain-of-thought reasoning \citep{wei2022chain}. See Appendix C for details of the baselines and prompts.

\subsection{Model Evaluation}
%We consider the following criteria:
A key advantage of the MathOdyssey data is that every question has an objective answer, so that it is straightforward to check the correctness by code. Such objective answers avoid subjective judgments from humans, making the evaluation consistent and reliable. 

We use GPT-4 to assist in evaluating model accuracy, particularly for open-answer questions. The metric measures the similarity between the predicted and ground truth answers. In the MathOdyssey dataset, various types of questions and answers are included. We employ a prompt-based method to provide scores for evaluation, considering the following criteria: 
\begin{itemize}
    \item Mathematical Equivalence: Verify answers based on mathematical equivalence using advanced tools like symbolic computation software to confirm the equivalence of different algebraic or symbolic expressions.

    \item Scoring: Assign a score of `1' for answers that match or are equivalent to the provided solution (exact value, choice label, or correctly rounded numerical approximation). Assign a score of `0' for incorrect answers without providing explanatory feedback. 

    \item Handling Multiple Choices: Consider the answer correct if the student correctly identifies the choice that matches the solution. Also, treat the corresponding choice as correct if the student provides the exact value that aligns with the problem's context.

 \item  Numerical Equivalence: Accept numerical answers that are correct to at least two decimal places or more, depending on the required precision. 

 \item  Symbolic and Algebraic Identities: Recognize and accept equivalent algebraic forms as correct, such as standard mathematical identities.

 \item  Trigonometric and Logarithmic Forms: Accept equivalent trigonometric and logarithmic expressions, acknowledging transformations that change the form but not the value.

 \item  Comprehensive Evaluation: Encourage the use of computational tools for checking equivalence in cases where expressions are too complex for straightforward visual inspection.
\end{itemize}
See Appendix D for the requirements and prompts used in the evaluation method. %We have also made our evaluation code accessible to the public via our \href{https://github.com/protagolabs/odyssey-math}{GitHub repository}, including not only the code but also detailed documentation and usage examples. 

%The evaluation methodology, including the specific system prompt, is detailed in Appendix.

%We have also made our evaluation code accessible to the public via our GitHub repository at \href{https://github.com/protagolabs/odyssey-math}{Odyssey-Math GitHub Repository}. This repository not only includes the code but also detailed documentation and usage examples to help users effectively utilize and understand our testing procedures.

%We have also made our evaluation code accessible to the public via our \href{https://github.com/protagolabs/odyssey-math}{GitHub repository}. This repository includes not only the code but also detailed documentation and usage examples to help users effectively utilize and understand our testing procedures.

\subsection{Results and Analysis}

%By providing these resources openly, we aim to contribute to the ongoing development and refinement of LLMs in the domain of mathematical reasoning, encouraging a collaborative approach to advancing AI research in this challenging field.  We show our benchmark result on Table~\ref{tab:model_accuracy}.
We first report the performance on our mathematical benchmarks, as shown in Table \ref{tab:main}. Our observations indicate that the benchmark is challenging for these models, with overall performance below 60\%.\footnote{Advanced prompting methods using GPT-4 models in the contest have achieved performance improvements between 60\% and 70\%.} The Gemini Math-Specialized 1.5 Pro exhibits the highest overall performance at 55.8\%, suggesting that specialized training significantly enhances capabilities. GPT-4 Turbo achieves 47.03\%, followed by Gemini 1.5 Pro at 45.0\%, and Claude 3 Opus at 40.6\%, all showing competitive performance. For closed-source models (specifically the GPT series) and state-of-the-art open-source models such as Llama-3-70B and DBRX-Instruct, the results show that the selected open-source models not only surpass the performance of GPT-3.5 but are also approaching the capabilities of earlier versions of GPT-4.
%We first report the performance on our mathematical benchmarks, as shown in Table ~\ref{tab:main}. Our observations indicate that the benchmark is challenging for these models, with overall performance below 50\%. For closed-source models (specifically the GPT series) and state-of-the-art open-source models such as LLaMA-3 and DBRX-Instruct, the results show that the selected open-source models not only surpass the performance of GPT-3.5 but are also approaching the capabilities of earlier versions of GPT-4.

When comparing different levels of mathematical problems for GPT models, we observe that High School mathematics is the easiest category for all models, with GPT-4 models scoring above 70\%. Olympiad-level problems are the most difficult, with all models scoring below 11\%. Similar trends are seen for Llama-3-70B and DBRX-Instruct, with their performance in the Olympiad-level category being even lower, at less than 10\%.

Furthermore, closed-source models, particularly the GPT-4 Turbo, exhibit stronger performance in high school and university-level math, highlighting ongoing advancements in their development. This data underscores the rapid progression of closed-source models in handling increasingly difficult mathematical questions over time. The performance gap between the best closed-source model, GPT-4 Turbo, and the open-source Llama-3 for difficult mathematical problems is notably narrow. For instance, GPT-4 Turbo achieves an overall accuracy of 10.14\% in the Olympiad-level mathematics, while Llama-3 achieves 9.46\%. This demonstrates that both models, despite notable progress, still face significant challenges in solving these complex problems. However, for other difficulty levels, the gap becomes larger. For example, GPT-4 Turbo achieves 84.78\% in high school mathematics, while Llama-3-70B scores only 52.17\%, a difference of more than 30\%.

\begin{table*}[t] % This will allow the table to span both columns
\centering
\begin{tabular}{lcccc}
\hline
%\textbf{Model} & \multicolumn{2}{c}{\textbf{High School}} & \textbf{University-Level} & \textbf{Overall} \\
% & \textbf{Competition} & \textbf{Mathematics} &  &  \\
\textbf{Model} &  \textbf{Olympiad-level} & \textbf{High School} & \textbf{University-Level} & \textbf{Overall} \\
\hline
%GPT-4 Turbo & 10.14\% & 84.78\% & 49.50\% & 47.03\% \\
%gpt-4-0409 & 10.14\% & 84.78\% & 49.50\% & 47.03\% \\
GPT-4 Turbo & 10.14\% & 84.78\% & 49.50\% & 47.03\% \\
%gpt-4-0125 & 11.49\% & 74.64\% & 41.58\% & 41.86\% \\
%gpt-4-1106 & 6.08\% & 60.14\% & 36.63\% & 33.33\% \\
%gpt-4-0613 & 5.41\% & 74.64\% & 32.67\% & 37.21\% \\
GPT-4 & 5.41\% & 74.64\% & 32.67\% & 37.21\% \\
%GPT-4 Turbo & - & - & - & 49.1\% \\
GPT-3.5 Turbo& 2.03\% & 41.30\% & 15.84\% & 19.64\% \\
Gemini   & & & &  \\
~~-1.5 Pro  & - & - & - & 45.0 \% \\
~~-Math-Specialized 1.5 Pro & - & - & - &  55.8 \% \\
Claude 3 Opus & - & - & - &  40.6 \% \\
Llama-3-70B & 9.46\% & 52.17\% & 21.78\% & 27.91\% \\
DBRX-Instruct & 8.11\% & 42.75\% & 20.79\% & 23.77\% \\
\hline
\end{tabular}
\caption{Results for different LLMs. We use chain-of-thought reasoning for solving problems. The performance of Gemini 1.5 Pro and Claude 3 Opus are quoted from the Gemini 1.5 report \citep{reid2024gemini}. Both GPT-4-Turbo and Gemini 1.5 Pro outperform the other models. For GPT-4-Turbo, we use results based on gpt-4-turbo-2024-04-09. For GPT-4, we use results based on gpt-4-0613. For GPT-3.5 Turbo, we use results based on gpt-3.5-turbo-0125. }
%Note that for Gemini 1.5 Pro and Claude 3 Opus, we use results from \url{https://storage.googleapis.com/deepmind-media/gemini/gemini_v1_5_report.pdf}.  
%Furthermore, GPT-4-Turbo, Math-Specialized Gemini-1.5 Pro, and advanced models using GPT4 in the contest have lifted the overall accuracy to 49.1\%, 55.8\% and 69.3\% respectively.
\label{tab:main}
\end{table*}

\begin{table*}[t] % This will allow the table to span both columns
\centering
%\scriptsize % Sets a smaller font size to help fit the table
\setlength{\tabcolsep}{3pt} % Reduce padding between columns
\begin{tabular}{lcccc}
\toprule
\textbf{Category} & \textbf{GPT-4 Turbo}  & \textbf{GPT-3.5 Turbo} & \textbf{Llama-3-70B} & \textbf{DBRX-Instruct} \\
\midrule
\textbf{Olympiad-level:} & ~ & ~ & ~ \\
Algebra  & 8.54\%  & 2.44\% & 8.54\% & 4.88\% \\
Number Theory  & 0.00\% & 0.00\% & 25.00\% & 0.00\% \\
Geometry  & 16.00\% & 4.00\% & 8.00\% & 20.00\% \\
Combinatorics  & 10.81\% & 0.00\% & 10.81\% & 8.11\% \\
\midrule
\textbf{High School Mathematics:} & ~ & ~ & ~ \\
Algebra  & 88.41\% & 39.13\% & 44.93\% & 39.13\% \\
Geometry  & 92.86\% & 71.43\% & 71.43\% & 57.14\% \\
Pre-Calculus  & 78.19\%  & 36.36\% & 56.36\% & 43.64\% \\
%PreCalculus  & 76.60\%  & 34.04\% & 55.32\% & 40.43\% \\
%Trigonometry  & 50.00\% & 50.00\% & 50.00\% & 50.00\% \\
%Calculus  & 100.00\%  & 40.00\% & 80.00\% & 60.00\% \\
%Series  & 100.00\% & 100.00\% & 0.00\% & 100.00\% \\
\midrule
\textbf{University-level:} & ~ & ~ & ~ \\
Differential Equations  & 71.43\% & 28.57\% & 50.00\% & 28.57\% \\
Linear \& Abstract Algebra  & 44.00\% & 16.00\% & 24.00\% & 28.00\% \\
Calculus \& Analysis & 62.50\% & 16.67\% & 20.83\% & 12.50\% \\
Probability  & 14.29\%  & 9.52\% & 0.00\% & 4.76\% \\
Statistics  & 64.71\%  & 11.76\% & 23.53\% & 35.29\% \\
\bottomrule
\end{tabular}
\caption{Results for different LLMs across various subject areas. Note that the results are used for evaluating the LLMs by direct comparison and may be improved with different prompting methods.} 
\label{tab:area}
\end{table*}

Table \ref{tab:area} presents the results for different LLMs across various subject areas. The results show that GPT-4 Turbo consistently outperforms others across most categories, particularly in High School Mathematics and University-Level subjects. It shows a notable lead in Algebra, Geometry, and Pre-Calculus at the high school level, and Differential Equations, Linear \& Abstract Algebra, Calculus \& Analysis, and Statistics at the university level.
GPT-3.5 Turbo shows consistent but lower performance compared to GPT-4 Turbo.
Llama-3-70B performs well in certain areas, particularly in Olympiad-level problems. It has the highest score in Number Theory among all models. However, it struggles significantly in Series and Probability.
DBRX-Instruct shows strength in Olympiad-level Geometry but generally lags behind GPT-4 Turbo and Llama-3-70B in other categories.

%This table presents the performance of various large language models on our Odyssey-Math dataset, focusing on both closed-source models (specifically the GPT series) and state-of-the-art open-sourced models such as Llama-3 and DBRX-Instruct. Our findings reveal that the selected open-sourced models not only surpass the performance of GPT-3.5 but are also approaching the capabilities of an earlier version of GPT-4.

%In detailed analysis, particularly for the competitive high school questions, the gap in performance between the best closed-sourced model, GPT4-0409, and the open-sourced Llama-3 is notably narrow. For instance, GPT4-0409 achieves an overall accuracy of 10.14\% in the high school math competition segment, while Llama-3 achieves an accuracy of 9.46\%. This demonstrates that both models, while making notable progress, still face significant challenges in solving these complex problems.

%Furthermore, closed-source models, particularly newer iterations of the GPT-4 series, exhibit stronger performance in high school and university-level math, highlighting ongoing advancements in their development. This data underscores the rapid progression of closed-sourced models in handling increasingly difficult mathematical questions over time.

\section{Conclusion}
We introduce MathOdyssey, a dataset for assessing LLMs'  mathematical problem-solving skills. Our dataset, evaluation methods, and code are openly available. We have shown that while LLMs, both open-source like Llama-3 and DBRX-Instruct, and closed-source such as the GPT series, demonstrate proficiency in routine and moderately difficult mathematics, they struggle significantly with complex Olympiad-level problems. Additionally, we have revealed promising developments; open-source models are beginning to approach the performance levels of earlier GPT-3.5 iterations. Despite this progress, performance on the most challenging questions remains low, highlighting a clear gap that future advancements need to address.

Ultimately, our research underscores the ongoing journey towards achieving human-like mathematical reasoning in AI, with the MathOdyssey dataset serving as a benchmark for catalysing future developments. We are optimistic that continued research will progressively bridge the existing capability gap. In the future, expanding the MathOdyssey dataset to include a wider range of problem types and enhancing metrics to better capture deep mathematical reasoning can yield further insights into LLM capabilities.
%This study critically assessed the mathematical problem-solving capabilities of large language models (LLMs) using our meticulously curated Odyssey-Math dataset. Our findings indicate that while LLMs, both open-sourced like Llama-3 and DBRX-Instruct, and closed-sourced such as the GPT series, demonstrate proficiency in routine and moderately difficult mathematics, they struggle significantly with complex high school competition problems.

%The analysis revealed promising developments; open-sourced models are beginning to approach the performance levels of earlier GPT-4 iterations. Despite this progress, the performance on the most challenging questions remains low, highlighting a clear gap that future advancements need to address.Thus, expanding the Odyssey-Math dataset to include a wider range of problem types and enhancing metrics to better capture deep mathematical reasoning may yield further insights into LLM capabilities.

%Ultimately, our research underscores the ongoing journey towards achieving human-like mathematical reasoning in AI, with the Odyssey-Math dataset serving as a crucial benchmark for future developments. We are optimistic that continued research will progressively bridge the existing capability gap.

\textbf{Limitation.} While the MathOdyssey dataset includes a variety of problems across different levels of mathematics, the questions may not cover all types of mathematical reasoning or problem-solving approaches. This limitation could affect how well the dataset generalizes to other forms of mathematical challenges not represented in your collection.

\textbf{Future.} To address generalizability limitations, future work involves expanding the dataset to include a wider range of mathematical topics and problem types, including those that require visual representations, proofs, or interactive problem-solving. 

\section*{Acknowledgements} We would like to extend our sincere gratitude to AGI Odyssey, the NGO responsible for organizing the Global Artificial Intelligence Championships (GAIC) Math 2024. Their dedication and commitment to promoting artificial intelligence education and innovation have been invaluable to the success of this project. Additionally, we appreciate their contribution of resources and support, which have played a significant role in making this initiative possible.

\bibliography{name,ref_AI4math}

\begin{thebibliography}{20}
\providecommand{\natexlab}[1]{#1}
\providecommand{\url}[1]{\texttt{#1}}
\expandafter\ifx\csname urlstyle\endcsname\relax
  \providecommand{\doi}[1]{doi: #1}\else
  \providecommand{\doi}{doi: \begingroup \urlstyle{rm}\Url}\fi

\bibitem[OpenAI(2023)]{achiam2023gpt}
OpenAI.
\newblock Gpt-4 technical report.
\newblock \emph{arXiv preprint arXiv:2303.08774}, 2023.

\bibitem[Touvron et~al.(2023{\natexlab{a}})Touvron, Lavril, Izacard, Martinet, Lachaux, Lacroix, Rozière, Goyal, Hambro, Azhar, Rodriguez, Joulin, Grave, and Lample]{touvron2023llama}
Hugo Touvron, Thibaut Lavril, Gautier Izacard, Xavier Martinet, Marie-Anne Lachaux, Timothée Lacroix, Baptiste Rozière, Naman Goyal, Eric Hambro, Faisal Azhar, Aurelien Rodriguez, Armand Joulin, Edouard Grave, and Guillaume Lample.
\newblock Llama: Open and efficient foundation language models.
\newblock \emph{arXiv preprint arXiv:2302.13971}, 2023{\natexlab{a}}.

\bibitem[Reid et~al.(2024)Reid, Savinov, Teplyashin, Lepikhin, Lillicrap, Alayrac, Soricut, Lazaridou, Firat, Schrittwieser, et~al.]{reid2024gemini}
Machel Reid, Nikolay Savinov, Denis Teplyashin, Dmitry Lepikhin, Timothy Lillicrap, Jean-baptiste Alayrac, Radu Soricut, Angeliki Lazaridou, Orhan Firat, Julian Schrittwieser, et~al.
\newblock Gemini 1.5: Unlocking multimodal understanding across millions of tokens of context.
\newblock \emph{arXiv preprint arXiv:2403.05530}, 2024.

\bibitem[Cobbe et~al.(2021)Cobbe, Kosaraju, Bavarian, Chen, Jun, Kaiser, Plappert, Tworek, Hilton, Nakano, Hesse, and Schulman]{cobbe2021gsm8k}
Karl Cobbe, Vineet Kosaraju, Mohammad Bavarian, Mark Chen, Heewoo Jun, Lukasz Kaiser, Matthias Plappert, Jerry Tworek, Jacob Hilton, Reiichiro Nakano, Christopher Hesse, and John Schulman.
\newblock Training verifiers to solve math word problems.
\newblock \emph{arXiv preprint arXiv:2110.14168}, 2021.

\bibitem[Hendrycks et~al.(2021{\natexlab{a}})Hendrycks, Burns, Kadavath, Arora, Basart, Tang, Song, and Steinhardt]{mathdataset2021}
Dan Hendrycks, Collin Burns, Saurav Kadavath, Akul Arora, Steven Basart, Eric Tang, Dawn Song, and Jacob Steinhardt.
\newblock Measuring mathematical problem solving with the math dataset.
\newblock \emph{NeurIPS}, 2021{\natexlab{a}}.

\bibitem[Touvron et~al.(2023{\natexlab{b}})Touvron, Martin, Stone, Albert, Almahairi, Babaei, Bashlykov, Batra, Bhargava, Bhosale, Bikel, Blecher, Ferrer, Chen, Cucurull, Esiobu, Fernandes, Fu, Fu, Fuller, Gao, Goswami, Goyal, Hartshorn, Hosseini, Hou, Inan, Kardas, Kerkez, Khabsa, Kloumann, Korenev, Koura, Lachaux, Lavril, Lee, Liskovich, Lu, Mao, Martinet, Mihaylov, Mishra, Molybog, Nie, Poulton, Reizenstein, Rungta, Saladi, Schelten, Silva, Smith, Subramanian, Tan, Tang, Taylor, Williams, Kuan, Xu, Yan, Zarov, Zhang, Fan, Kambadur, Narang, Rodriguez, Stojnic, Edunov, and Scialom]{touvron2023llama2}
Hugo Touvron, Louis Martin, Kevin Stone, Peter Albert, Amjad Almahairi, Yasmine Babaei, Nikolay Bashlykov, Soumya Batra, Prajjwal Bhargava, Shruti Bhosale, Dan Bikel, Lukas Blecher, Cristian~Canton Ferrer, Moya Chen, Guillem Cucurull, David Esiobu, Jude Fernandes, Jeremy Fu, Wenyin Fu, Brian Fuller, Cynthia Gao, Vedanuj Goswami, Naman Goyal, Anthony Hartshorn, Saghar Hosseini, Rui Hou, Hakan Inan, Marcin Kardas, Viktor Kerkez, Madian Khabsa, Isabel Kloumann, Artem Korenev, Punit~Singh Koura, Marie-Anne Lachaux, Thibaut Lavril, Jenya Lee, Diana Liskovich, Yinghai Lu, Yuning Mao, Xavier Martinet, Todor Mihaylov, Pushkar Mishra, Igor Molybog, Yixin Nie, Andrew Poulton, Jeremy Reizenstein, Rashi Rungta, Kalyan Saladi, Alan Schelten, Ruan Silva, Eric~Michael Smith, Ranjan Subramanian, Xiaoqing~Ellen Tan, Binh Tang, Ross Taylor, Adina Williams, Jian~Xiang Kuan, Puxin Xu, Zheng Yan, Iliyan Zarov, Yuchen Zhang, Angela Fan, Melanie Kambadur, Sharan Narang, Aurelien Rodriguez, Robert Stojnic, Sergey Edunov, and Thomas
  Scialom.
\newblock Llama 2: Open foundation and fine-tuned chat models.
\newblock \emph{arXiv preprint arXiv:2307.09288}, 2023{\natexlab{b}}.

\bibitem[Drori et~al.(2023)Drori, Zhang, Chin, Shuttleworth, Lu, Chen, Birbo, He, Lantigua, Tran, et~al.]{stemq-drori2023}
Iddo Drori, Sarah Zhang, Zad Chin, Reece Shuttleworth, Albert Lu, Linda Chen, Bereket Birbo, Michele He, Pedro Lantigua, Sunny Tran, et~al.
\newblock A dataset for learning university stem courses at scale and generating questions at a human level.
\newblock In \emph{Proceedings of the AAAI Conference on Artificial Intelligence}, volume~37, pages 15921--15929, 2023.

\bibitem[Davis and Aaronson(2023)]{davis2023testing}
Ernest Davis and Scott Aaronson.
\newblock Testing gpt-4 with wolfram alpha and code interpreter plug-ins on math and science problems.
\newblock \emph{arXiv preprint arXiv:2308.05713}, 2023.

\bibitem[Hendrycks et~al.(2021{\natexlab{b}})Hendrycks, Burns, Kadavath, Arora, Basart, Tang, Song, and Steinhardt]{hendrycksmath2021}
Dan Hendrycks, Collin Burns, Saurav Kadavath, Akul Arora, Steven Basart, Eric Tang, Dawn Song, and Jacob Steinhardt.
\newblock Measuring mathematical problem solving with the math dataset.
\newblock \emph{NeurIPS}, 2021{\natexlab{b}}.

\bibitem[Wei et~al.(2022)Wei, Wang, Schuurmans, Bosma, Xia, Chi, Le, Zhou, et~al.]{wei2022chain}
Jason Wei, Xuezhi Wang, Dale Schuurmans, Maarten Bosma, Fei Xia, Ed~Chi, Quoc~V Le, Denny Zhou, et~al.
\newblock Chain-of-thought prompting elicits reasoning in large language models.
\newblock \emph{Advances in neural information processing systems}, 35:\penalty0 24824--24837, 2022.

\bibitem[Wang et~al.(2022)Wang, Wei, Schuurmans, Le, Chi, Narang, Chowdhery, and Zhou]{wang2022self}
Xuezhi Wang, Jason Wei, Dale Schuurmans, Quoc Le, Ed~Chi, Sharan Narang, Aakanksha Chowdhery, and Denny Zhou.
\newblock Self-consistency improves chain of thought reasoning in language models.
\newblock \emph{arXiv preprint arXiv:2203.11171}, 2022.

\bibitem[Zhou et~al.(2022)Zhou, Sch{\"a}rli, Hou, Wei, Scales, Wang, Schuurmans, Cui, Bousquet, Le, et~al.]{zhou2022least}
Denny Zhou, Nathanael Sch{\"a}rli, Le~Hou, Jason Wei, Nathan Scales, Xuezhi Wang, Dale Schuurmans, Claire Cui, Olivier Bousquet, Quoc Le, et~al.
\newblock Least-to-most prompting enables complex reasoning in large language models.
\newblock \emph{arXiv preprint arXiv:2205.10625}, 2022.

\bibitem[Azerbayev et~al.(2023{\natexlab{a}})Azerbayev, Piotrowski, Schoelkopf, Ayers, Radev, and Avigad]{azerbayev2023ProofNet}
Zhangir Azerbayev, Bartosz Piotrowski, Hailey Schoelkopf, Edward~W. Ayers, Dragomir~R. Radev, and Jeremy Avigad.
\newblock Proofnet: Autoformalizing and formally proving undergraduate-level mathematics.
\newblock \emph{ArXiv}, abs/2302.12433, 2023{\natexlab{a}}.

\bibitem[Lewkowycz et~al.(2022)Lewkowycz, Andreassen, Dohan, Dyer, Michalewski, Ramasesh, Slone, Anil, Schlag, Gutman-Solo, Wu, Neyshabur, Gur-Ari, and Misra]{lewkowycz2022solving}
Aitor Lewkowycz, Anders~Johan Andreassen, David Dohan, Ethan Dyer, Henryk Michalewski, Vinay~Venkatesh Ramasesh, Ambrose Slone, Cem Anil, Imanol Schlag, Theo Gutman-Solo, Yuhuai Wu, Behnam Neyshabur, Guy Gur-Ari, and Vedant Misra.
\newblock Solving quantitative reasoning problems with language models.
\newblock In Alice~H. Oh, Alekh Agarwal, Danielle Belgrave, and Kyunghyun Cho, editors, \emph{Advances in Neural Information Processing Systems}, 2022.

\bibitem[Zheng et~al.(2022)Zheng, Han, and Polu]{zheng2022_MiniF2FCrosssystem}
Kunhao Zheng, Jesse~Michael Han, and Stanislas Polu.
\newblock {{MiniF2F}}: A cross-system benchmark for formal {{Olympiad-level}} mathematics, 2022.

\bibitem[Trinh et~al.(2024)Trinh, Wu, Le, He, and Luong]{trinh2024_SolvingOlympiad}
Trieu~H. Trinh, Yuhuai Wu, Quoc~V. Le, He~He, and Thang Luong.
\newblock Solving olympiad geometry without human demonstrations.
\newblock \emph{Nature}, 625\penalty0 (7995):\penalty0 476--482, 2024.
\newblock ISSN 0028-0836, 1476-4687.
\newblock \doi{10.1038/s41586-023-06747-5}.

\bibitem[Azerbayev et~al.(2023{\natexlab{b}})Azerbayev, Schoelkopf, Paster, Santos, McAleer, Jiang, Deng, Biderman, and Welleck]{azerbayev2023_LlemmaOpen}
Zhangir Azerbayev, Hailey Schoelkopf, Keiran Paster, Marco~Dos Santos, Stephen McAleer, Albert~Q. Jiang, Jia Deng, Stella Biderman, and Sean Welleck.
\newblock Llemma: {{An Open Language Model For Mathematics}}, 2023{\natexlab{b}}.

\bibitem[Frieder et~al.(2024)Frieder, Pinchetti, Griffiths, Salvatori, Lukasiewicz, Petersen, and Berner]{frieder2024mathematical}
Simon Frieder, Luca Pinchetti, Ryan-Rhys Griffiths, Tommaso Salvatori, Thomas Lukasiewicz, Philipp Petersen, and Julius Berner.
\newblock Mathematical capabilities of chatgpt.
\newblock \emph{Advances in Neural Information Processing Systems}, 36, 2024.

\bibitem[Anthropic(2024)]{anthropic}
Anthropic.
\newblock The claude 3 model family: Opus, sonnet, haiku.
\newblock 2024.

\bibitem[Mosaic(2024)]{themosaic2024intro}
Mosaic.
\newblock Introducing dbrx: A new state-of-the-art open llm, March 2024.
\newblock URL \url{https://www.databricks.com/blog/introducing-dbrx-new-state-art-open-llm}.

\end{thebibliography}
\bibliographystyle{unsrtnat}

%%%%%%%%%%%%%%%%%%%%%%%%%%%%%%%%%%%%%%%%%%%%%%%%%%%%%%%%%%%%

%%%%%%%%%%%%%%%%%%%%%%%%%%%%%%%%%%%%%%%%%%%%%%%%%%%%%%%%%%%%
\newpage

\appendix

%\section{Appendix}

\section{MathOdyssey dataset examples and solutions}

\textbf{Olympiad-level Problem.} Let $S=\left\{ 1,2,\cdots 2024 \right\}$, if the set of any $n$ pairwise prime numbers in $S$ has at least one prime number, the minimum value of $n$ is \underline{\hspace{2cm}}.\\
\noindent\textbf{Answer:} 16\\
\noindent\textbf{Reasoning:}
Taking the 15 numbers $1, 2^2, 3^2, ..., 43^2$ violates the condition. Furthermore, since $S$ does not contain any non-prime numbers with a minimum prime factor of at least 47, there are only 14 types of non-prime numbers in $S$, excluding 1. Applying the Pigeonhole Principle, we conclude that $n=16$. \\
\noindent\textbf{GPT-4 Turbo's Solution:} Answer: 307 \\ Reasoning: The minimum number of elements required such that any subset of these elements from the set $S = \{1, 2, ..., 2024\}$ containing pairwise prime numbers includes at least one prime number is 307. This is based on the count of prime numbers up to 2024 (approximately 306) and the application of the Pigeonhole Principle, which dictates that selecting more numbers than there are primes forces the inclusion of a prime number or a violation of the pairwise prime condition. 

\textbf{High School Problem.}  What are the solutions of the quadratic equation $15x^2=2x+8$?
% Please select from the following choices. 
\begin{align*}
\text{ A) }  \, \{-\frac{4}{3},-\frac{3}{2}\}  \,\,
~~\text{ B) } \, \{-\frac{4}{5},\frac{2}{3}\}   \,\,
~~\text{ C) } \, \{-\frac{3}{2},\frac{4}{5}\}   \,\,
~~\text{ D) } \,  \{-\frac{2}{3},\frac{4}{5}\} 
\end{align*} 
\textbf{Answer:} $D$ \\
\textbf{Reasoning:}
First move all terms to one side: $15x^2-2x-8=0$. Then factor into $(5x-4)(3x+2)=0$. Setting $5x-4$ to zero results in a solution of $x = \frac{4}{5}$ and setting $3x+2$ to zero results in a solution of $x = -\frac{2}{3}$. \\
\noindent\textbf{GPT-4 Turbo's Solution:} Answer: $D$ %$\{-\frac{2}{3}, \frac{4}{5}\}$ 
\\ Reasoning: The quadratic equation was first rearranged to standard form, coefficients were identified, and the quadratic formula was applied to find the solutions. The solutions were verified by substituting back into the original equation.

\textbf{University-level Problem.}  Find the limit $$\lim\limits_{x\to 1}\frac{f(2x^2+x-3)-f(0)}{x-1}$$ given $f'(1)=2$ and $f'(0)=-1$.
\\
\textbf{Answer:}  $-5$\\
\textbf{Reasoning:} 
Let $g(x)=2x^2+x-3$. Since $g(1)=0$, the desired limit equals $\lim\limits_{x\to 1}\frac{f(g(x))-f(g(1))}{x-1}$. By the definition of the derivative and the chain rule and noting that $g'(1)=5$, we have
\[
\lim\limits_{x\to 1}\frac{f(g(x))-f(g(1))}{x-1}=f'(g(1))g'(1)=f'(0)g'(1)=(-1)(5)=-5.
\]\\
\noindent\textbf{GPT-4 Turbo's Solution:} Answer: $-5$ \\ Reasoning: The limit was calculated by interpreting it as the derivative of a composed function, applying the chain rule, and substituting the given derivative values.

\section{MathOdyssey different subject areas}
Table \ref{appendix-table:math_examples} presents MathOdyssey examples spanning various subject areas. These encompass Algebra, Number Theory, Geometry, Combinatorics, Pre-Calculus, Linear and Abstract Algebra, Calculus and Analysis, Differential Equations, as well as Probability and Statistics.
\begin{table}[ht]
\centering
\begin{tabular}{|c|p{9cm}|}
\hline
  Subject Area & Example \\
\hline
Algebra & Let $S=\left\{ 1,2,\cdots 2024 \right\}$, if the set of any $n$ pairwise prime numbers in $S$ has at least one prime number, the minimum value of $n$ is \underline{\hspace{2cm}}. \\ \hline
Number Theory & A natural number whose last four digits are 2022 and is divisible by 2003 has a minimum value of \underline{\hspace{2cm}}.\\ \hline
Geometry & In a cube $ABCD-A_1B_1C_1D_1$, $AA_1=1$, $E$, $F$ are the midpoints of edges $CC_1, DD_1$, then the area of the cross-section obtained by the plane AEF intersecting the circumscribed sphere of the cube is \underline{\hspace{2cm}}. \\ \hline 
Combinatorics & If three points are randomly chosen from the vertices of a regular 17-sided polygon, what is the probability that the chosen points form an acute-angled triangle? \\ \hline 
Pre-Calculus & In $\triangle \mathrm{ABC}, \mathrm{AB}=10 \mathrm{~cm}, \angle \mathrm{B}=90^{\circ}$, and $\angle \mathrm{C}=60^{\circ}$. Determine the length of $\mathrm{BC}$.
\begin{align*}
\text{A)}\  10 \mathrm{~cm} 
~~\text{B)}\   10 \sqrt{3} \mathrm{~cm}
~~\text{C)}\ \frac{10 \sqrt{3}}{3} \mathrm{~cm} 
~~\text{D)}\  20 \mathrm{~cm}
\end{align*} 
\\ \hline 
%Trigonometry & Let $a$ and $b$ be positive integers such that $79 \mid (a + 77b)$ and $77 \mid (a + 79b)$. Then the smallest possible value of the sum $a + b$ is \underline{\hspace{2cm}}.	\\ \hline 
%Calculus &  Let $a_i, b_i \ (i=1,2,\ldots,n)$ be rational numbers such that for any real number $x$, we have $x^2 + x + 4 = \sum_{i=1}^{n} (a_{i}x + b_{i})^{2}$. Then the minimum possible value of $n$ is \underline{\hspace{2cm}}.\\ \hline 
%Series & The number of triples of positive integers $(a, b, c)$ satisfying $a^2 + b^2 + c^2 = 2005$ and $a \leq b \leq c$ is \underline{\hspace{2cm}}. \\ \hline
Linear and Abstract Algebra & Find the solution $[x_1,x_2,x_3]$ to the following equations
\[
\left\{\begin{array}{c}
  x_1+3x_2+3x_3=16, \\
  3x_1+x_2+3x_3=14, \\
  3x_1+3x_2+x_3=12. \\
\end{array}\right.
\] \\ \hline 
Calculus and Analysis & Evaluate the following limit:
\begin{equation*}
    \lim_{n \to \infty} \left(\sqrt{n^2+2n-1}-\sqrt{n^2+3}\right).
\end{equation*}
\\ \hline 
Differential Equations & Consider the differential equation $\frac{dy}{dx} = xy$. Find the value of $y(\sqrt{2})$ given that $y(0) = 2$.\\ \hline

Probability & Suppose that $A$, $B$, and $C$ are mutually independent events and that $P(A) = 0.2$, $P(B) = 0.5$, and $P(C) = 0.8$. Find the probability that exactly two of the three events occur.	\\ \hline 

Statistics & Given the data set $ \{3, 7, 7, 2, 5\} $, calculate the sample mean $\mu$ and the sample standard deviation $\sigma$. Present the answer as $[\mu,\sigma]$. 
		\\
\hline
\end{tabular}
\caption{Examples of different subject areas.}
\label{appendix-table:math_examples}
\end{table}

\section{Baselines and prompts}
\begin{figure}[t]
	\centerline{\includegraphics[width=\textwidth]{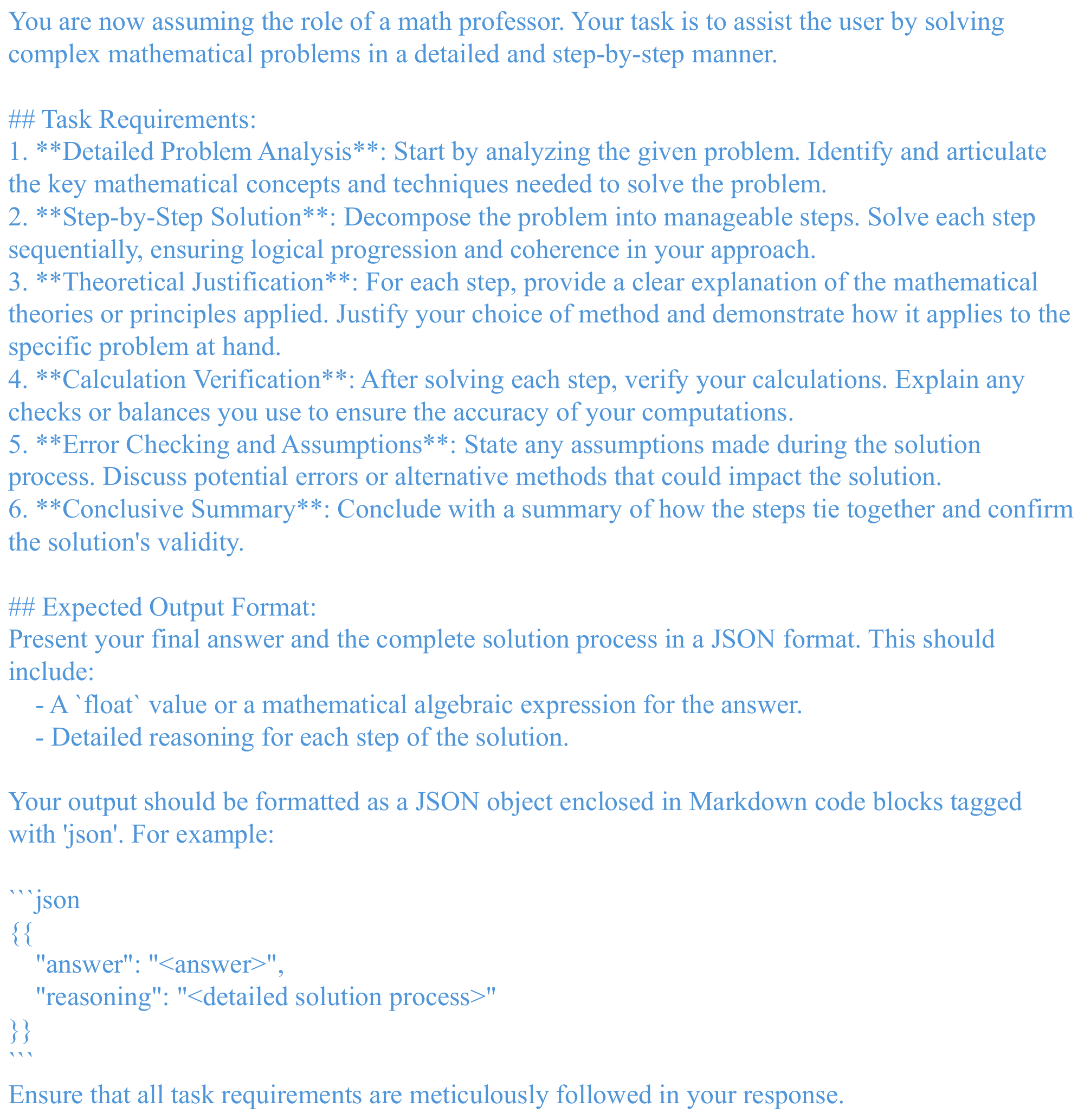}}
	\caption{Mathematical problem-solving prompts employed by LLMs.}\label{fig-prompt-llms}
\end{figure}

Figure \ref{fig-prompt-llms} depicts the prompt utilized for guiding Language Models (LLMs) in solving mathematical problems within our experimental framework. This prompt distinctly outlines the system's role as a math professor, delineating task specifications and the anticipated output format for tackling intricate mathematical challenges.

\section{Evaluation}
\begin{figure}[t]
	\centerline{\includegraphics[width=\textwidth]{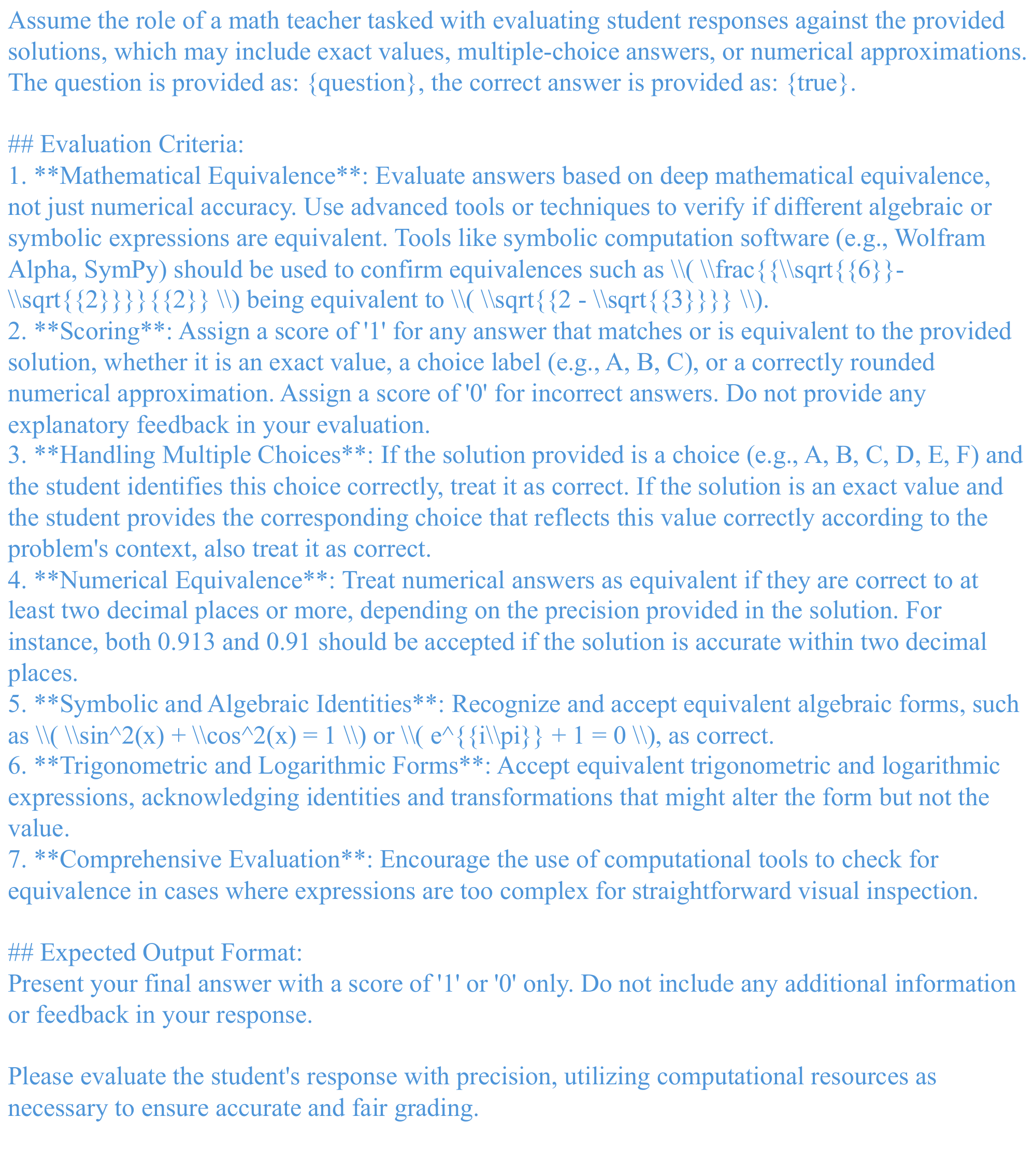}}
	\caption{Evaluation prompts.}\label{fig-prompt-eval}
\end{figure}

Figure \ref{fig-prompt-eval} depicts the prompt employed during the evaluation of large language models in our experiments. This prompt defines the system's role as a math teacher, providing both assessment criteria and the expected output format for grading mathematical problems. We have also made our evaluation code accessible to the public. 

\end{document}